\documentclass{article}
\usepackage{hyperref}
\usepackage{breakurl}
\usepackage[dvips]{graphicx,color}
\usepackage{paralist}
\usepackage{amsmath}
\usepackage[letterpaper, left=1cm, right=1.25cm, top=1cm, bottom=1.25cm]{geometry}
\usepackage{subfigure}
\usepackage{multirow}
\usepackage{fancyhdr}

\title{An Improved Objective Evaluation Measure for Border Detection in Dermoscopy Images}

\author{%
        M. Emre Celebi \footnote{Corresponding author email: \href{mailto:ecelebi@lsus.edu}{ecelebi@lsus.edu}}\\
        Dept. of Computer Science\\Louisiana State Univ., Shreveport, LA, USA\\
\and 
        Gerald Schaefer\\
        School of Engineering and Applied Science\\Aston Univ., Birmingham, UK\\
\and
        Hitoshi Iyatomi\\
        Dept. of Electrical Informatics\\Hosei Univ., Tokyo, Japan\\
\and 
        William V.\ Stoecker\\
        Stoecker \& Associates, Rolla, MO, USA\\
\and
        Joseph M.\ Malters\\
        The Dermatology Center, Rolla, MO, USA\\
\and
        James M.\ Grichnik\\
        Dept.\ of Medicine\\Duke Univ.\ Medical Center, Durham, NC, USA\\
       }
\begin{document}
\maketitle

\begin{abstract}

\textbf{Background:} Dermoscopy is one of the major imaging modalities used in the diagnosis of melanoma and other pigmented skin lesions. Due to the difficulty and subjectivity of human interpretation, dermoscopy image analysis has become an important research area. One of the most important steps in dermoscopy image analysis is the automated detection of lesion borders. Although numerous methods have been developed for the detection of lesion borders, very few studies were comprehensive in the evaluation of their results. \textbf{Methods:} In this paper, we evaluate five recent border detection methods on a set of 90 dermoscopy images using three sets of dermatologist-drawn borders as the ground-truth. In contrast to previous work, we utilize an objective measure, the Normalized Probabilistic Rand Index, which takes into account the variations in the ground-truth images. \textbf{Conclusion:} The results demonstrate that the differences between four of the evaluated border detection methods are in fact smaller than those predicted by the commonly used XOR measure.

\end{abstract}

\section{Introduction}
\label{sec_intro}

Invasive and in-situ malignant melanoma together comprise one of the most rapidly increasing cancers in the world. Invasive melanoma alone has an estimated incidence of 62,480 and an estimated total of 8,420 deaths in the United States in 2008~\cite{Jemal08}. Early diagnosis is particularly important since melanoma can be cured with a simple excision if detected early.
\par
Dermoscopy, also known as epiluminescence microscopy, is a non-invasive skin imaging technique that uses optical magnification and either liquid immersion and low angle-of-incidence lighting or cross-polarized lighting, making subsurface structures more easily visible when compared to conventional clinical images~\cite{Argenziano02}. Dermoscopy allows the identification of dozens of morphological features such as pigment network, dots/globules, streaks, blue-white areas, and blotches~\cite{Menzies03}. This reduces screening errors, and provides greater differentiation between difficult lesions such as pigmented Spitz nevi and small, clinically equivocal lesions~\cite{Steiner93}. However, it has been demonstrated that dermoscopy may actually lower the diagnostic accuracy in the hands of inexperienced dermatologists~\cite{Binder95}. Therefore, in order to minimize the diagnostic errors that result from the difficulty and subjectivity of visual interpretation, the development of computerized image analysis techniques is of paramount importance~\cite{Fleming98}.
\par
Automated border detection is often the first step in the automated or semi-automated analysis of dermoscopy images~\cite{Celebi07a}. It is crucial for the image analysis for two main reasons. First, the border structure provides important information for accurate diagnosis as many clinical features such as asymmetry, border irregularity, and abrupt border cutoff are calculated directly from the border. Second, the extraction of other important clinical features such as atypical pigment network~\cite{Fleming98}, globules~\cite{Stoecker05}, and blue-white areas~\cite{Celebi08b} critically depends on the accuracy of border detection. Automated border detection is a challenging task due to several reasons:
\begin{inparaenum}[(i)]
\item low contrast between the lesion and the surrounding skin,
\item irregular and fuzzy lesion borders,
\item artifacts and intrinsic cutaneous features such as black frames, skin lines, blood vessels, hairs, and air bubbles,
\item variegated coloring inside the lesion, and
\item fragmentation due to various reasons such as scar-like depigmentation.
\end{inparaenum}

\par
Numerous methods have been developed for border detection in dermoscopy images~\cite{Celebi09}. Recent approaches include fuzzy c-means clustering~\cite{Schmid99a,Cucchiara02,Zhou09}, gradient vector flow snakes~\cite{Erkol05}, thresholding followed by region growing~\cite{Iyatomi06,Iyatomi08}, meanshift clustering~\cite{Melli06}, color quantization followed by spatial segmentation~\cite{Celebi07b}, statistical region merging~\cite{Celebi08a}, two-stage k-means++ clustering followed by region merging~\cite{Zhou08}, and contrast enhancement followed by k-means clustering~\cite{Delgado08}. Some of these studies used subjective visual examination to evaluate their results. Others used objective measures including Hance \emph{et al.}'s XOR measure~\cite{Hance96}, sensitivity \& specificity, precision \& recall, error probability, and pixel misclassification probability~\cite{Guillod02}. These measures require borders drawn by dermatologists, which serve as the ground truth. In this paper, we refer to the computer-detected borders as \emph{automatic borders} and those determined by dermatologists as \emph{manual borders}.
\par
In a recent study, Guillod \emph{et al.}~\cite{Guillod02} demonstrated that a single dermatologist, even one who is experienced in dermoscopy, cannot be used as an absolute reference for evaluating border detection accuracy. In addition, they emphasized that manual borders are not precise, with inter-dermatologist borders and even intra-dermatologist borders showing significant disagreement, so that a probabilistic model of the border is preferred to an absolute gold-standard model.
\par
Only a few of the above-mentioned studies used borders determined by multiple dermatologists. Guillod \emph{et al.}~\cite{Guillod02} used fifteen sets of borders determined by five dermatologists over a minimum period of one month. They constructed a probability image for each lesion by associating a misclassification probability with each pixel based on the number of times it was selected as part of the lesion. The automatic borders were then compared against these probability images. Iyatomi \emph{et al.}~\cite{Iyatomi06,Iyatomi08} modified Guillod \emph{et al.}'s method by combining the manual borders that correspond to each image into one using the majority vote rule. The automatic borders were then compared against these combined ground-truth images. Celebi \emph{et al.}~\cite{Celebi08a} compared each automatic border against multiple manual borders independently.
\par
In this paper, we evaluate the performance of five recent automated border detection methods on a set of 90 dermoscopy images using three sets of manual borders as the ground-truth. In contrast to prior studies, we employ an objective criterion that takes into account the variations in the ground-truth images.
\par
The rest of the paper is organized as follows. Section~\ref{sec_rev} reviews the objective measures used previously in the border detection literature. Section~\ref{sec_new_meas} describes a recent measure that takes into account the variations in the ground-truth images. Section~\ref{sec_exp} presents the experimental setup and discusses the results obtained, while Section~\ref{sec_conc} concludes the paper.

\section{Review of Objective Measures for Border Detection Evaluation}
\label{sec_rev}

All of the objective measures mentioned in Section \ref{sec_intro}, except for Guillod \emph{et al.}'s probabilistic measure, are based on the concepts of true/false positive/negative defined in Table~\ref{tab_def}. For example, if a lesion pixel is detected as part of the background skin, this pixel is considered to be a False Negative. On the other hand, if a background pixel is detected as part of the lesion, it is considered as a False Positive. Note that in the remainder of this paper, True Positive (TP), False Negative (FN), False Positive (FP), and True Negative (TN) will refer to the number of pixels that satisfy these criteria.

\begin{table}[!ht]
\centering
\caption{ \label{tab_def} Definitions of true/false positive/negative. `Actual' and `detected' pixels refer to a pixel in the manual border and the corresponding pixel in the automatic border, respectively. }
\begin{tabular}{cccc}
 \multicolumn{2}{c}{} & \multicolumn{2}{c}{Detected Pixel} \\ 
\cline{3-4}
\multicolumn{2}{c|}{Actual Pixel}  & Lesion & Background \\ 
\hline
\multicolumn{2}{c|}{Lesion} & True Positive (TP) & False Negative (FN) \\ 
\multicolumn{2}{c|}{Background} & False Positive (FP) & True Negative (TN) \\ 
\end{tabular}
\end{table}

\subsection{XOR Measure}

The XOR measure, first used by Hance \emph{et al.}~\cite{Hance96} quantifies the percentage border detection error as
\begin{equation}
\label{eq_xor}
\begin{array}{l}
 \mbox{Error} = \frac{{\mbox{Area}\left( {\mbox{AB}\, \oplus \,\mbox{MB}} \right)}}{{\mbox{Area}\left( \mbox{MB} \right)}} \times 100\%  \\ 
 \quad \quad \,\,\,\, = \frac{\mbox{FP} + \mbox{FN}}{\mbox{TP} + \mbox{FN}} \times 100\%  \\ 
 \end{array}
\end{equation}
where \emph{AB} and \emph{MB} are the binary images obtained by filling the automatic and manual borders, respectively, $\oplus$ is the exclusive-OR (XOR) operation that gives the pixels for which \emph{AB} and \emph{MB} disagree, and Area($I$) denotes the number of pixels in the binary image $I$. The drawback of this composite measure is that it tends to favor larger lesions due to the size term in the denominator.

\subsection{Sensitivity \& Specificity}
Sensitivity (true positive rate) and specificity (true negative rate) are commonly used evaluation measures in medical studies. In our application domain, the former corresponds to the percentage of correctly detected lesion pixels, whereas the latter corresponds to the percentage of correctly detected background pixels. Mathematically, these measures are given by
\begin{equation}
\begin{array}{l}
 \mbox{Sensitivity} = \frac{\mbox{TP}}{\mbox{TP} + \mbox{FN}} \times 100\%  \\ 
 \mbox{Specificity} \, = \frac{\mbox{TN}}{\mbox{FP} + \mbox{TN}} \times 100\%  \\ 
 \end{array}
\label{equ_SESP}
\end{equation}

Note that an automatic border that encloses the corresponding manual border will have a perfect ($100\%$) sensitivity. On the other hand, an automatic border border that is completely enclosed by the corresponding manual border will have a perfect specificity. Therefore, it is crucial not to interpret these measures in isolation from each other.

\subsection{Precision \& Recall}
Precision (positive predictive value) and recall are commonly used evaluation measures in information retrieval studies. Precision refers to  the percentage of correctly detected lesion pixels over all the pixels detected as part of the lesion and is defined as
\begin{equation}
\mbox{Precision} = \frac{\mbox{TP}}{\mbox{TP} + \mbox{FP}} \times 100\%
\end{equation}

Recall is equivalent to sensitivity as defined in \eqref{equ_SESP}. Note that as in the case of sensitivity and specificity, precision and recall measures should be interpreted together.

\subsection{Error Probability}
Error probability refers to the percentage of pixels incorrectly detected as part of the lesion or background over all the pixels. It is calculated as
\begin{equation}
\mbox{Error probability} = \frac{\mbox{FP} + \mbox{FN}}{\mbox{TP} + \mbox{FN} + \mbox{FP} + \mbox{TN}} \times 100\%
\end{equation}

The drawback of this composite measure is that it disregards the distributions of the classes. For example, consider a small lesion of size $20,000$ pixels in a large image of size $768 \times 512$ pixels. An automatic border of size $40,000$ pixels that encloses the manual border for this lesion will have an error probability of about $5\%$ despite the fact that the automatic border is twice as large as the manual border.

\subsection{Pixel Misclassification Probability}
In~\cite{Guillod02} the probability of misclassification for a pixel $(i,j)$ is defined as
\begin{equation}
p(i,j) = 1 - \frac{n(i,j)}{N}
\end{equation}
where $N$ is the number of observations (manual + automatic borders), and $n(i,j)$ is the number of times pixel $(i,j)$ was selected as part of the lesion. For each automatic border, the detection error is given by the mean probability of misclassification over the pixels inside the border
\begin{equation}
 \mbox{Error} = \frac{\sum\limits_{(i,j) \in AB} {p(i,j)}} {\mbox{TP} + \mbox{FP}} \times 100\%
\end{equation}

\subsection{Error Measures Used in Previous Studies}
Table~\ref{tab_eval} compares recent border detection methods based on their evaluation methodology: the number of human experts who determined the manual borders, the number of images used in the evaluations (and the diagnostic distribution of these images if available), and the measure used to quantify the border detection error. It can be seen that:

\begin{itemize}
	\item Recent studies used objective measures to validate their results, whereas earlier studies relied on visual assessment.
	\item Only 5 out of 19 studies involve more than one expert in the evaluation of their results.
	\item XOR measure is the most commonly used objective error function despite the fact that it is not trivial to extend this measure to
capture the variations in multiple manual borders. 
\end{itemize}

\begin{table}
\centering
\caption{ \label{tab_eval} Evaluation of border detection methods (b: benign, m: melanoma) }
\begin{tabular}{c|c|c|c|c}
\hline 
Ref.\ & Year & \# Experts & \# Images (Distribution) & Error Measure\ (Value) \\
\hline
\hline
\cite{Zhou09} & 2009 & 1 & 100 (70 b / 30 m) & Sens.\ (78\%) \& Spec.\ (99\%) \\
\hline
\cite{Celebi08a} & 2008 & 3 & 90 (65 b / 25 m) & XOR (10.63\%)\\
\hline
\cite{Zhou08} & 2008 & 1 & 67 & XOR (14.63\%) \\
\hline
\cite{Delgado08} & 2008 & 1 & 100 (70 b / 30 m) & XOR (2.73\%)\\
\hline
\cite{Mendonca07} & 2007 & 1 & 50 & Error probability (16\%)\\
\hline
\cite{Mendonca07} & 2007 & 1 & 50 & Error probability (21\%)\\
\hline
\cite{Celebi07b} & 2007 & 2 & 100 (70 b / 30 m) & XOR (12.02\%)\\
\hline
\cite{Iyatomi06} & 2006 & 5 & 319 (244 b / 75 m) & Prec.\ (94.1\%) \& Rec.\ (95.2\%)\\
\hline
\cite{Melli06} & 2006 & nr & 117 & Sens.\ (95\%) \& Spec.\ (96\%)\\ 
\hline
\cite{Erkol05} & 2005 & 2 & 100 (70 b / 30 m) & XOR (15.59\%)\\
\hline
\cite{Galda03} & 2003 & 0 & nr & nr\\
\hline
\cite{Cucchiara02} & 2002 & 0 & 600 & Visual\\
\hline
\cite{Hintz-Madsen01} & 2001 & 0 & nr & nr\\
\hline
\cite{Haeghen00} & 2000 & 5 & 30 & Visual\\
\hline
\cite{Donadey00} & 2000 & 1 & 30 & Visual\\
\hline
\cite{Schmid99a} & 1999 & 1 & 400 & Visual\\
\hline
\cite{Schmid99b} & 1999 & 1 & 300 & Visual\\
\hline
\cite{Gao98} & 1998 & 1 & 57 & XOR (36.50\%)\\
\hline
\cite{Gao98} & 1998 & 1 & 57 & XOR (24.71\%)\\
\hline 
\end{tabular}
\end{table}

\section{Proposed Measure for Border Detection Evaluation}
\label{sec_new_meas}

The objective measures reviewed in the previous section share a common deficiency. They do not take into account the variations in the manual borders. Given an automatic border, the XOR measure, sensitivity \& specificity, precision \& recall, and error probability can only be defined with respect to a single manual border. Therefore, it is not possible to use these measures with multiple manual borders. Although the methods described in~\cite{Guillod02},~\cite{Iyatomi06,Iyatomi08}, and~\cite{Celebi08a} allow the use of multiple manual borders; these methods do not accurately capture the variations in the manual borders. For example, using Guillod \emph{et al.}'s measure an automated border that is entirely enclosed by the manual borders would get a very low error. Iyatomi \emph{et al.}'s method discounts the variation in the manual borders by simple majority voting, while Celebi \emph{et al.}'s approach does not produce a scalar error value, which makes comparisons more difficult.
\par
In this paper we propose to use a recent, more elaborate probabilistic measure, namely the Normalized Probabilistic Rand Index (NPRI)~\cite{Unnikrishnan07} to evaluate border detection accuracy. We first describe the Probabilistic Rand Index (PRI) \cite{Unnikrishnan05}. Consider a set of manual segmentations $\left\{ {S_1, \ldots ,S_K } \right\}$ of an image $X = \left\{ {x_1, \ldots ,x_N } \right\}$
  consisting of $N$ pixels. Let $S_{test}$ be the segmentation that is to be compared with the manually labeled set. We denote the label of point $x_i$ by $l_i ^{S_{test} }$ in segmentation $S_{test}$ and by $l_i ^{S_k }$ in the manually segmented image $S_k$.
\par
The motivation behind the PRI is that a segmentation is judged as `good' if it correctly identifies the pairwise relationships between the pixels as defined in the ground truth segmentations. In addition, a proper segmentation quality measure should penalize inconsistencies between the test and ground-truth label pair relationships proportionally to the level of consistency between the ground-truth label pair relationships. Based on this, the PRI is defined as
\begin{equation}
\label{equ_pri}
 \mbox{PRI}\left( {S_{test} ,\left\{ {S_k } \right\}} \right) = \frac{{\sum\nolimits_{\scriptstyle i,j \hfill \atop 
  \scriptstyle i < j \hfill} {c_{ij} p_{ij}  + (1 - c_{ij} )(1 - p_{ij} )} }}{{\left( \begin{array}{c}
 N \\ 
 2
 \end{array} \right)}}\,
\end{equation}
where $I(.)$ is a boolean function defined as
\begin{equation*}
I(t) = \left\{ \begin{gathered}
  1\quad t = \text{true} \hfill \\
  0\quad t = \text{false} \hfill \\ 
\end{gathered}  \right.
\end{equation*}

$c_{ij} \in \{0,1\}$ denotes the event of a pair of pixels $x_i$ and $x_j$ having the same label in the test image $S_{test}$
\begin{equation}
 c_{ij}  = {\rm I}\left( {l_i ^{S_{test} }  = l_j ^{S_{test} } } \right)
\end{equation}

Note that the denominator in \eqref{equ_pri} denotes the number of possible distinct pixel pairs.
Given the $K$ manually labeled images, we can compute the empirical probability of the label relationship of a pixel pair $x_i$ and $x_j$ by
\begin{equation}
p_{ij}  = \frac{1}{K}\sum\limits_{k = 1}^K {{\rm I}\left( {l_i ^{S_k }  = l_j ^{S_k } } \right)} 
\end{equation}

The PRI is always within the interval $[0,1]$, and an index of 0 or 1 can only be achieved when all of the ground-truth segmentations agree or disagree on every pixel pair relationship. A score of 0 indicates that every pixel pair in the test image has the opposite relationship as every pair in the ground-truth segmentations, while a score of 1 indicates that every pixel pair in the test image has the same relationship as every pair in the ground-truth segmentations.
\par
The PRI has one disadvantage. Although the index values are in $[0,1]$, there is no expected value for a given segmentation. That is, it is impossible to know if any given score is good or bad. In addition, the score of a segmentation of one image cannot be compared with the score of a segmentation of another image. The Normalized Probabilistic Rand Index (NPRI) addresses this drawback by normalizing the PRI as follows
\begin{equation}
\mbox{Normalized Index} = \frac{{\mbox{Index} - \mbox{Exp.\ Index}}}{{\mbox{Max.\ Index} - \mbox{Exp.\ Index}}}
\end{equation}

The maximum index is taken as 1 while the expected value of the index is calculated as
\begin{equation}
E\left[ {\mbox{PRI}\left( {S_{test} ,\left\{ {S_k } \right\}} \right)} \right] = \frac{{\sum\nolimits_{\scriptstyle i,j \hfill \atop 
  \scriptstyle i < j \hfill} {p'_{ij} p_{ij}  + (1 - p'_{ij} )(1 - p_{ij} )} }}{{\left( \begin{array}{l}
 N \\ 
 2 \\ 
 \end{array} \right)}} 
\end{equation}

Let $\Phi$ be the number of images in the entire data set, and $K_{\phi}$ be the number of ground-truth segmentations of image $\phi$. Then $p'_{ij}$ can be expressed as
\begin{equation}
p'_{ij}  = E\left[ {c_{ij} } \right] = \frac{1}{\Phi }\sum\limits_\phi  {\frac{1}{{K_\phi  }}} \sum\limits_{k = 1}^{K_\phi  } {{\rm I}\left( {l_i ^{S^\phi  _k }  = l_j ^{S^\phi  _k } } \right)} 
\end{equation}

Since in the computation of the expected values no assumptions are made with regards to the number or size of regions in the segmentation, and all of the ground-truth data is used, the NPR indices are comparable across images and segmentations. 

\section{Experimental Results and Discussion}
\label{sec_exp}

The proposed evaluation method was tested on a set of 90 dermoscopy images (23 invasive malignant melanoma and 67 benign) obtained from the EDRA Interactive Atlas of Dermoscopy~\cite{Argenziano02}, and three private dermatology practices \cite{Celebi08a}. The benign lesions included nevocellular nevi and dysplastic nevi.
\par
Manual borders were obtained by selecting a number of points on the lesion border, connecting these points by a second-order B-spline and finally filling the resulting closed curve. Three sets of manual borders were determined by dermatologists Dr. William Stoecker (WS), Dr. Joseph Malters (JM), and Dr. James Grichnik (JG) using this method.
\par
Five recent automated border detection methods were included in the experiments. These were orientation-sensitive fuzzy c-means method (OSFCM)~\cite{Schmid99a}, dermatologist-like tumor extraction algorithm (DTEA)~\cite{Iyatomi06,Iyatomi08}, meanshift clustering method (MS)~\cite{Melli06}, modified JSEG method (JSEG)~\cite{Celebi07b}, and the statistical region merging method (SRM)~\cite{Celebi08a}. Table~\ref{tab_xor} gives the mean and standard deviation errors as evaluated by the commonly used XOR measure (\ref{eq_xor}). The best results, i.e.\ the lowest mean errors, in each row are shown in \textbf{bold}.

\begin{table}
\centering
\caption{ \label{tab_xor} XOR measure statistics: mean (standard deviation) }
\begin{tabular}{c|c|c|c|c|c|c}
Dermatologist & Diagnosis & OSFCM & DTEA & MS & JSEG & SRM \\
\hline
\hline
\multirow{6}{*}{WS} & \multirow{2}{*}{Benign} & {22.995} & {\bf{10.513}} & {11.527} & {10.832} & {11.384} \\ 
 &  & {(12.614)} & {(4.728)} & {(9.737)} & {(6.359)} & {(6.232)} \\ 
 & \multirow{2}{*}{Melanoma} & {28.311} & {11.853} & {13.292} & {13.745} & {\bf{10.294}} \\ 
 &  & {(15.245)} & {(5.998)} & {(7.418)} & {(7.590)} & {(5.838)} \\ 
 & \multirow{2}{*}{All} & {24.354} & {\bf{10.855}} & {11.978} & {11.577} & {11.106} \\ 
 &  & {(13.449)} & {(5.081)} & {(9.193)} & {(6.772)} & {(6.120)} \\
\hline 
\multirow{6}{*}{JM} & \multirow{2}{*}{Benign} & {25.535} & {10.367} & {10.802} & {10.816} & {\bf{10.186}} \\ 
 &  & {(11.734)} & {(3.771)} & {(6.332)} & {(5.227)} & {(5.683)} \\ 
 & \multirow{2}{*}{Melanoma} & {26.743} & {10.874} & {12.592} & {12.981} & {\bf{10.500}} \\ 
 &  & {(14.508)} & {(5.016)} & {(7.202)} & {(6.316)} & {(8.137)} \\ 
 & \multirow{2}{*}{All} & {25.843} & {10.496} & {11.259} & {11.370} & {\bf{10.266}} \\ 
 &  & {(12.426)} & {(4.101)} & {(6.571)} & {(5.570)} & {(6.351)} \\
\hline
\multirow{6}{*}{JG} & \multirow{2}{*}{Benign} & {27.506} & {12.091} & {12.224} & {12.257} & {\bf{10.561}} \\ 
 &  & {(12.789)} & {(5.220)} & {(7.393)} & {(6.588)} & {(5.152)} \\ 
 & \multirow{2}{*}{Melanoma} & {27.574} & {12.675} & {12.168} & {13.414} & {\bf{10.411}} \\ 
 &  & {(15.836)} & {(6.865)} & {(7.479)} & {(7.379)} & {(5.860)} \\ 
 & \multirow{2}{*}{All} & {27.523} & {12.240} & {12.210} & {12.553} & {\bf{10.523}} \\ 
 &  & {(13.538)} & {(5.650)} & {(7.373)} & {(6.775)} & {(5.308)} \\ 
\end{tabular}
\end{table}

It can be seen that the results vary significantly across the border sets, highlighting the subjectivity of human experts in the border determination procedure. Overall, the SRM method achieves the lowest mean errors followed by the DTEA and JSEG methods. It should be noted that, with the exception of SRM, the error rates increase in the melanoma group which is possibly due to the presence of higher border irregularity and color variation in these lesions. With respect to consistency, the best methods are DTEA followed by the SRM and JSEG methods.
\par
Table~\ref{tab_npri} shows the border detection quality statistics as evaluated by the proposed NPRI measure. Note that, in this table, higher mean values indicate lower border detection errors, whereas higher standard deviation values indicate lower consistency, respectively.

\begin{table}
\centering
\caption{ \label{tab_npri} NPRI measure statistics: mean (standard deviation) }
\begin{tabular}{c|c|c|c|c|c}
Diagnosis & OSFCM & DTEA & MS & JSEG & SRM \\
\hline
\hline
\multirow{2}{*}{Benign} & 0.520 & \bf{0.785} & 0.774 & 0.775 & \bf{0.785} \\ 
 & {(0.247)} & {(0.079)} & {(0.137)} & {(0.114)} & {(0.109)} \\
\hline
\multirow{2}{*}{Melanoma} & {0.520} & {0.783} & {0.762} & {0.748} & {\bf{0.811}} \\ 
 & {(0.258)} & {(0.108)} & {(0.161)} & {(0.141)} & {(0.092)} \\ 
\hline
\multirow{2}{*}{All} & {0.520} & {0.784} & {0.771} & {0.768} & {\bf{0.791}} \\ 
  & {(0.248)} & {(0.087)} & {(0.142)} & {(0.122)} & {(0.105)} \\
\hline
\end{tabular}
\end{table}

It can be seen that the ranking remains the same: SRM and DTEA are still the most accurate and consistent methods. However, using the NPRI measure, the differences between the methods have become smaller. In addition, this measure considers the variations in the manual borders simultaneously and produces a scalar value, which makes comparisons among methods much easier.
\par
\begin{figure}[!ht]
\centering
\includegraphics[totalheight=0.36\textwidth,draft=false]{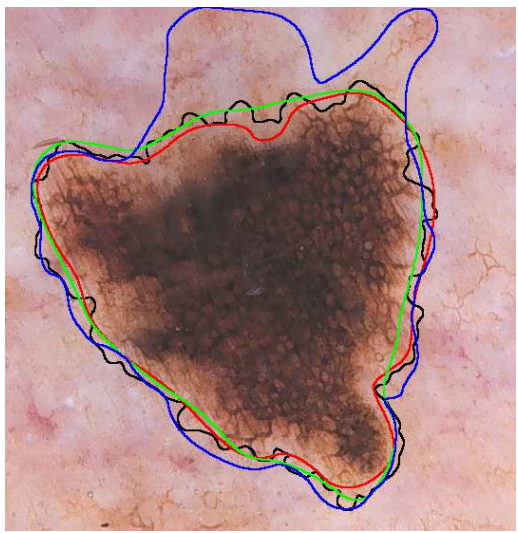}
\caption{ \label{fig_sample} Sample border detection result }
\end{figure}

Figure~\ref{fig_sample} illustrates one advantage of using the NPRI measure. Here the manual borders are shown in red, green, and blue, whereas the border determined by the DTEA method is shown in black. The border detection errors with respect to the red, green, and blue borders calculated using the XOR measure are 10.872\%, 9.342\%, and 20.958\%, respectively. It can be concluded that, with respect to the first two dermatologists, the DTEA method has an average accuracy (see Table~\ref{tab_xor}). On the other hand, with respect to the third dermatologist, the automatic method is quite inaccurate. The NPRI value in this case is 0.814, which is above the average over the entire data set (see Table~\ref{tab_npri}). This was expected, since this measure does not penalize the automatic border in those regions where dermatologist agreement is low.

\section{Conclusions and Future}
\label{sec_conc}

In this paper, we evaluated five recent automated border detection methods on a set of 90 dermoscopy images using three sets of manual borders as ground-truth. We proposed the use of an objective measure, the Normalized Probabilistic Rand Index, which takes into account variations in the ground-truth. The results demonstrated that the differences between four of the evaluated border detection methods were in fact smaller than those predicted by the commonly used XOR measure. Future work will be directed towards the expansion of the image set and the inclusion of more dermatologists in the evaluations.

\section*{Acknowledgments}

This publication was made possible by grants from The Louisiana Board of Regents (LEQSF2008-11-RD-A-12) and The National Institutes of Health (SBIR \#2R44 CA-101639-02A2). The assistance of Joseph M.\ Malters, M.D., and James M. Grichnik, M.D.\ in obtaining the manual borders is gratefully acknowledged.

\end{document}